%% file: main.tex
\documentclass[letterpaper, 10 pt, conference]{ieeeconf}  

\IEEEoverridecommandlockouts                              

\usepackage{times}

\usepackage{multicol}
\usepackage{graphicx}
\usepackage{subcaption}
\usepackage{array}
\usepackage{booktabs}

\usepackage{graphics} 
\usepackage{epsfig} 
\usepackage{mathptmx} 
\usepackage{times} 
\usepackage{amsmath} 
\usepackage{amssymb}  
\usepackage{sty_cls_files/Definitions}
\usepackage{xcolor}

\title{\LARGE \bf
Deep Forward and Inverse Perceptual Models for Tracking and Prediction
}

\author{ \authorblockN{ Alexander Lambert, Amirreza Shaban, Amit Raj, Zhen Liu and Byron Boots}
\thanks{Alexander Lambert and Byron Boots are affiliated with
the Institute for Robotics and Intelligent Machines, Georgia Institute
of Technology, USA. \texttt{alambert6@gatech.edu, bboots@cc.gatech.edu}. Amirezza Shaban, Amit Raj, and Zhen Liu are with the College of Computing, Georgia Institute of Technology, USA. \texttt{\{amirreza, amit.raj, liuzhen1994\}@gatech.edu} }}

\begin{document}

\maketitle
\thispagestyle{empty}
\pagestyle{empty}


\vspace{-0.4in}
\begin{abstract}
We consider the problems of learning forward models that map state to high-dimensional images and inverse models that map high-dimensional images to state in robotics. 
Specifically, we present a perceptual model for generating video frames from state with deep networks, and provide a framework for its use in tracking and prediction tasks. We show that our proposed model greatly outperforms standard deconvolutional methods and GANs for image generation, producing clear, photo-realistic images. We also  develop a convolutional neural network model for state estimation and compare the result to an Extended Kalman Filter to estimate robot trajectories.  We validate all models on a real robotic system. 

\end{abstract}


\input{intro}

\input{obs_model}

\input{results}

\input{conclusions}




\bibliographystyle{IEEEtran}
\bibliography{IEEEabrv,bibfile}

\end{document}

%% file: intro.tex
\section{Introduction \& Related Work}

Several fundamental problems in robotics, including state estimation, prediction, and motion planning rely on accurate models that can map state to measurements (forward models) or measurements to state (inverse models). Classic examples include the measurement models for global positioning systems, inertial measurement units, or beam sensors that are frequently used in simultaneous localization and mapping~\cite{thrun2005probabilistic}, or the forward and inverse kinematic models that map joint configurations to workspace and \emph{vice-versa}. Some of these models can be very difficult to derive analytically, and, in these cases, roboticists have often resorted to machine learning to infer accurate models directly from data. For example, complex nonlinear forward kinematics have been modeled with techniques as diverse as Bayesian networks \cite{dearden2005learning} and Bezier Splines~\cite{ulbrich2012learning}, and many researchers have tackled the problem of learning inverse kinematics with nonparametric methods like locally weighted projection regression (LWPR)~\cite{vijayakumar2000locally,d2001learning}, mixtures of experts~\cite{damas2012online}, and Gaussian Process Regression~\cite{nguyen2009local}. While these techniques are able to learn accurate models, they rely heavily on prior knowledge about the kinematic relationship between the robot state-space and work-space.


Despite the important role that forward and inverse models have played in robotics, there has been little progress in defining these models for very high-dimensional sensor data like images and video. This has been disappointing: cameras are a cheap, reliable source of information about the robot and its environment, but the precise relationship between a robot pose or configuration, the environment, and the generated image is extremely complex. A possible solution to this problem is to \emph{learn} a forward model that directly maps the robot pose or configuration to high-dimensional perceptual space  or an inverse model that maps new images to the robot pose or configuration. Given these models, one can directly and accurately solve a wide range of common robotics problems including recursive state estimation, sequential prediction, and motion planning. 

In this work, we explore the idea of directly learning forward and inverse perceptual models that relate high-dimensional images and low-dimensional robot state. Specifically, we use  deep neural networks to learn both forward and inverse perceptual models for a camera pointed at the manipulation space of a Barret WAM arm. While recent work on convolutional neural networks (CNNs) provides a fairly straightforward framework for learning inverse models that can map images to robot configurations, learning accurate generative (forward) models remains a challenge. 

Generative neural networks have recently shown much promise in addressing the problem of mapping low-dimensional encodings to a high-dimensional pixel-space \cite{dosovitskiy2015flownet,kulkarni2015deep,goodfellow2014,radford}. The generative capacity of these approaches is heavily dependent on learning a strictly parametric model to map input vectors to images. Using deconvolutional networks for learning controllable, kinematic transformations of objects has previously been demonstrated, as in \cite{dosovitskiy2016,tatarchenko2015}. However, these models have difficulty reproducing clear images with matching textures, and have mainly been investigated on affine transformations of simulated objects.

Learning to predict frames has also been conducted on two-dimensional robot manipulation tasks. Finn et al.~\cite{finn2016unsupervised} propose an LSTM-based network to predict next-frame images, given the current frame and state-action pair. In order to model pixel transformations, the authors make use of composited convolutions with either unconstrained or affine kernels. The generated image frames appear to reproduce linear motion in the scene, but also appear to have difficulty replicating multi-degree-of-freedom dynamics. Given that forward prediction is conducted by recursive input of predicted frames, error compounds during sequential prediction and prediction quality quickly degrades over future timesteps. 

An alternative approach to generating images directly, after applying transformations in a low-dimensional encoding, is to learn a transformation in the high-dimensional output space. One can then re-use pixel information from observed images to reconstruct new views from the same scene. This has been proposed in previous studies~\cite{ganin2016deepwarp,efros2016}. The authors in~\cite{efros2016}, for instance, learn a model to generate a flow-field transformation from an input image-pose pair derived from synthetic data. This is subsequently applied to a reference frame, effectively rotating the original image to a previously unseen viewpoint. Using confidence masks to combine multiple flow-fields generated from different reference frames is also proposed. However, these frames are selected randomly from the training data. 

In this work, we propose deep forward and inverse perceptual models for a camera pointed at the manipulation space of a Barrett WAM arm. The forward model maps a 4-dimensional arm configuration to a $\left(256\times256\times3\right)$-dimensional RGB image and the inverse model maps $\left(256\times256\times3\right)$-dimensional images to 4-dimensional configurations. A major contribution of this work is a new forward model that extends the image-warping model in \cite{efros2016} with a \emph{non-parametric} reference-frame selection component. We show that this model can generate sharp, \emph{near photo-realistic}, images from a never-before-seen 4-dimensional arm configurations and greatly outperform state-of-the-art deconvolutional networks~\cite{dosovitskiy2016,tatarchenko2015}. We also demonstrate how the forward model can also be used for predicting occlusions. This may be desirable in certain robot tasks, where maintaining visibility of the robot is required for monitoring execution, for instance. Lastly, we show how to design a convolutional inverse model, and use the two models in tandem for tracking the configuration of the robot from an unknown initial configuration and prediction to future never-before-seen images. We perform several experiments validating our approach, both in simulation and on a real robot. 

%% file: obs_model.tex
\section{Sensor Model Design}
\vspace{-1.0mm}

\begin{figure*}[h]
    \begin{center}
        \includegraphics[width=0.60\textwidth]{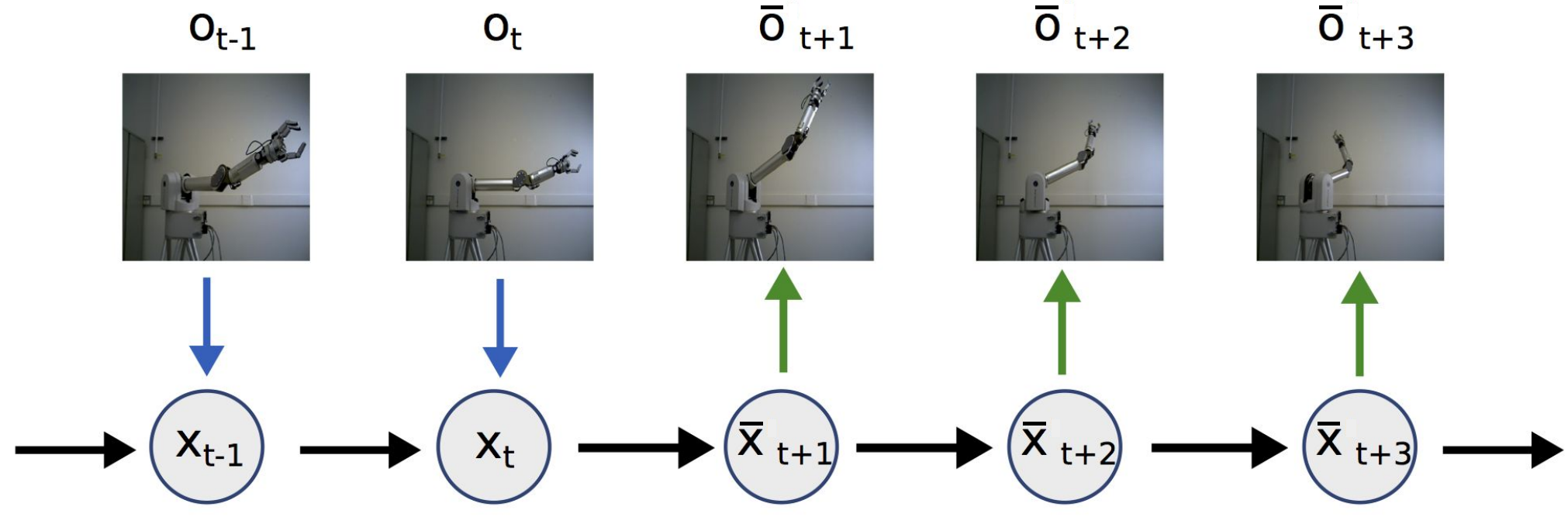}
        \label{placeholder}
    \end{center}
    \caption{\small The proposed framework makes use of an inverse sensor model (blue) that can be used to infer the state $x_i$ from observation $o_i$. Once the robot state has been estimated, a motion plan can be generated given the known state space and system dynamics. The resulting trajectory can then be mapped into image space using a forward sensor model (green), generating a high-fidelity video of what the robot expects to see while executing the motion plan. }\vspace{-4.0mm}
    \label{fig:high_level_framework}
\end{figure*}

Tracking and prediction are important tasks in robotics. 
Consider the problem of tracking state and predicting future images, illustrated in Fig.\ref{fig:high_level_framework}. Given a history of  $O_t = \left( o_i \right)_{i=0}^t$ where $o \in \mathbb{R}^m$, we wish to track the state of the system $x \in \mathbb{R}^n$ for the corresponding sequence $X_t=\left( x_i \right)_{i=0}^t$  This can be accomplished using an \emph{inverse model} that maps images to state. Conversely, given a  sequence of expected states $\overline{X}_t=  \left( \overline{x}_i \right)_{i=t+1}^T$ (which may be obtained from motion-planning), each state $\bar{x}_i$  can be mapped to a corresponding observation prediction $\bar{o}_i$ by a \emph{forward model}.

The tasks of mapping the configuration-space joint values to (observed) robot poses, and \textit{vice-versa}, is similar to the problem of forward/inverse kinematics for modeling robot manipulator mechanics. 
Traditionally, such mechanical models are integrated with visual information using separately-constructed image-projection models, for techniques such as Position-based Visual Servoing \cite{chaumette2006visual,corke1993visual}. In the present context, however, the kinematics and projection relations are represented jointly by the learned forward/inverse models.



\subsection{A Forward Sensor Model}
\label{fwd_model}
\begin{figure*}[h]
    \centering
    \begin{subfigure}[b]{0.9\textwidth}
    	\centering
        \includegraphics[width=\textwidth]{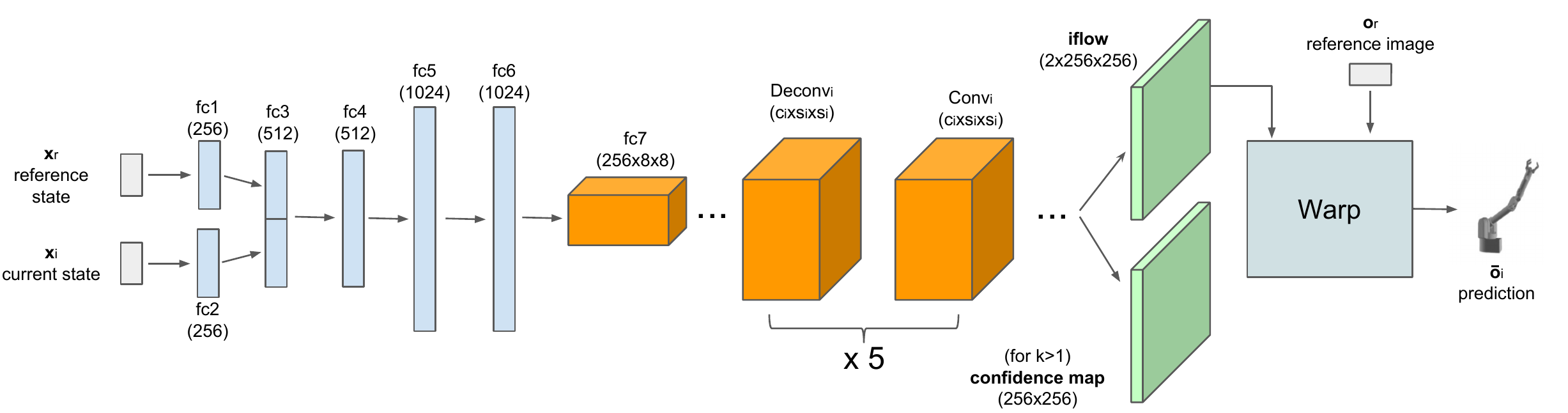}
        \caption{\small Forward branch}
        \label{fig:arch_param}
    \end{subfigure}
    \begin{subfigure}[b]{0.7\textwidth}
    	\centering
        \includegraphics[width=\textwidth]{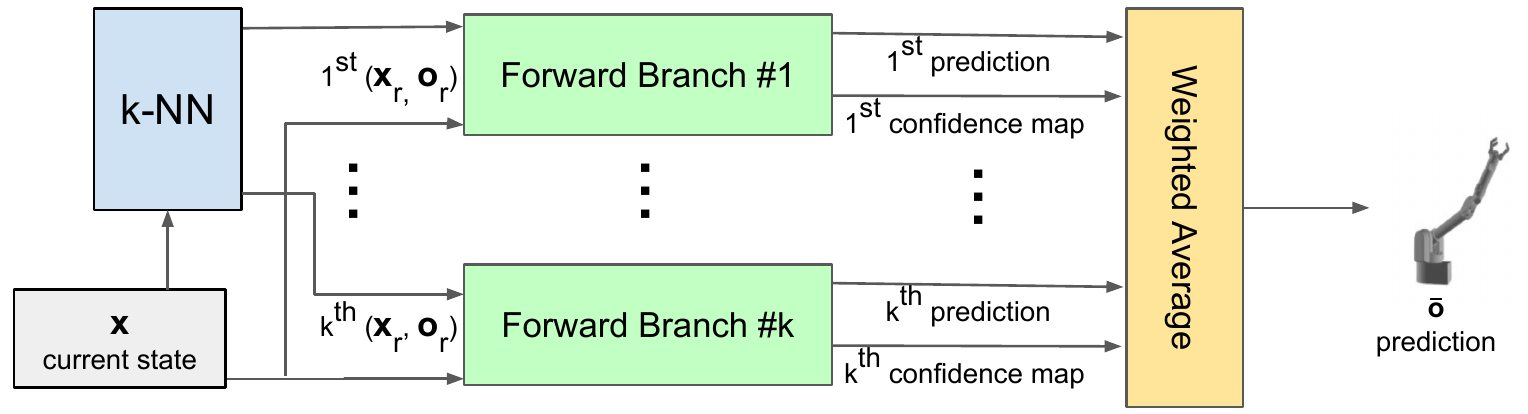}
        \caption{\small $k$-NN-FLOW model}
        \label{fig:arch_kNN}
    \end{subfigure}
    \caption{\small{Forward sensor model architecture. 
    \protect(\subref{fig:arch_param}) Single parametric branch example with flow-field output (iflow). \protect(\subref{fig:arch_kNN}) Complete $k$-NN-FLOW model architecture. $k$-Nearest neighbor (image, state)-pairs selected by the $k$-NN module, and passed to individual parametric branches (having shared weights). Resulting warped images are weighed by their corresponding confidence maps and summed together to produce final output image. Refer to main text for details.}} \vspace{-3.0mm}
    \label{fig:arch}
\end{figure*}

We use a generative observation model $g$ to perform a mapping from joint-space values $x_i$ to pixel-space values $o_i$ for a trajectory sequence of future states $\overline{X}_t$. 
Instead of training a neural network to predict the RGB values completely from scratch, our network is comprised of two parts: 1) a \emph{non-parametric} component, where given a state value $x_i$, a reference pair $(x_r, o_r)$ is found such that $o_r$ has a similar appearance as the output image $o_i$; and 2) a parametric component, where given the same state value $x_i$ and the reference state-image pair $(x_r, o_r)$, we learn to predict the pixel flow field $\overrightarrow{h}(x_i, x_r)$ from reference image $o_r$ to the image $o_i$ (using image-warping layers similar to\cite{efros2016}).
The final prediction is made by warping the reference image $o_r$ with the predicted flow field $\overrightarrow{h}$:
\begin{equation}
g(x_i) = \mathrm{warp}(o_r, \overrightarrow{h}). 
\end{equation}
As long as there is a high correlation between visual appearance of the reference image $o_r$ and the output image $o_i$, warping the reference image results in much higher-fidelity predictions when compared to models which map input directly to RGB values \cite{tatarchenko2015}. The overall architecture is shown in Fig.\ref{fig:arch}.



We use a nearest neighbor algorithm in the configuration space to find $o_r$. More precisely, we take a subset of the training set $\Fcal_p \subseteq \Fcal$, consisting of configuration-image pairs and store the data in a KD-tree \cite{maneewongvatana1999s}. We can quickly find a reference pair by searching for the training data pair closest to the current joint configuration $x_i$ in configuration space. Since the state-space is low-dimensional we can find this nearest neighbor very efficiently. The 1-nearest neighbor algorithm, despite its simplicity, has been successfully used in many computer vision applications~\cite{benetis2002nearest, guillaumin2009tagprop}, and can pick an accurate $(x_r, o_r)$ pair from the training set. 

In practice, the output of Nearest Neighbor 
is limited to the elements inside the training set and cannot generalize to unseen data very well. To address this problem, in the second part of our model, we train a deep neural network to warp the predicted $o_r$ (with the joint configuration $x_r$) to the target observation $o_i$ (with the joint configuration $x_i$). A diagram of this parametric model is shown in Figure \ref{fig:arch_param}. Given a current and reference state vector as input, the network is trained to produce a flow tensor to warp the reference image (similarly to \cite{efros2016}). Intuitively, the Nearest Neighbor component 
(Figure \ref{fig:arch_kNN}) provides the RGB values for a nearby joint configuration, and the neural network learns to move the pixels to produce an image which corresponds to the target joint vector. The input is first mapped to a high dimensional space using six fully-connected layers. The resulting vector is reshaped into a tensor with size $256 \times 8 \times 8$. The spatial resolution is increased from $8\times8$ to $2^8\times2^8$ by using $5$ deconvolution layers. Following the deconvolutions with convolutional layers was found to qualitatively improve the output images. Finally, we use the predicted flow field of size $2\times256\times256$ to warp the reference image. To allow end-to-end training, we used a differentiable image sampling method with a bi-linear interpolation kernel in the warping layer~\cite{jaderberg2015spatial}.

Using $L_2$ prediction loss, the final optimization function is defined as follows:
\begin{equation}
 \min_{W} L(W) = \sum_i || o_i - \mathrm{warp}(o_r, {\overrightarrow{h}}_{W}(x_i, x_r)) ||^2 + \lambda ||W||^2
\end{equation}
in which $W$  contains the network parameters and $\lambda$ is the regularization coefficient. 

To prevent over-fitting, we randomly sample one of 10-nearest neighbors in the training phase. 
Training on the forward model was performed end-to-end using the Caffe library~\cite{jia2014caffe}, which we extended to support the warping layer. ADAM ~\cite{kingma2014adam} was used to optimize the network. 



\subsubsection{Forward Model with $k$-Nearest Neighbors}

Using a reference image works well if all visible portions of the arm in $o_i$ are \emph{also} visible in the reference image $o_r$ \cite{efros2016}. However, portions of the arm may be visible in some reference images and not others. 
As such, we could improve the prediction fidelity by warping more than one reference image, and merging the results. The idea of warping a single nearest neighbor can be extended to warping an ensemble of $k$-nearest neighbor reference images, with each neighbor contributing separately to the prediction.

In addition to the flow field, each network in the ensemble also predicts a $256\times256$ confidence map (as shown in Fig.\ref{fig:arch}). We use these confidence maps to compute the weighted sum of different predictions in the ensemble and compute the final prediction. We refer to this general, multi-neighbor formulation of the forward model as $k$-NN-FLOW. 

\subsubsection{Prediction using the Forward Model}

Given a trajectory or sequence of states $X_t= \left( x_i \right)_{i=t+1}^T$, it may be desirable to obtain the corresponding observations $O_t= \left( o_i \right)_{i=t+1}^T$. For instance, an expected trajectory $\overline{X}_t$ could be obtained by using a motion-planning algorithm in simulation. The resulting solution would then be translated into observation space using the forward sensor model, generating a predictive video containing photo-realistic image frames. Generating each image $o_i$ is accomplished simply by performing a forward pass on each corresponding state value $x_i$ using the architecture described in Section \ref{fwd_model}.

\subsubsection{Tracking using the Forward Model}
\label{fwd_tracking}
A common approach to state estimation is to use a Bayesian filter with a generative sensor model to provide a correction conditioned on measurement uncertainty. For instance, the forward model described in Section \ref{fwd_model} could be used to provide such a state update, and potentially allow for a single model to be used for both tracking and prediction. We therefore consider this approach as a suitable baseline for comparison to the inverse-forward framework that we ultimately suggest (see Section~\ref{inv_model} below). 

To track a belief-state distribution, we derive an Extended Kalman Filter from the forward sensor model. For the parametric component, this is a straightforward process, as deep neural networks are inherently amenable to linearization. To perform the correction step, the Jacobian $J$ is computed as the product of the layer-wise Jacobians:
\begin{equation}
J  = J^{(L)} \times J^{(L - 1)} \times \dots \times J^{(0)}
\end{equation}
where $L$ is the total number of layers in the network branch.
The dimensionality of the observations makes it impractical to compute the Kalman Gain $\left(K = \Sigma J^T(J\Sigma J^T + R)^{-1}\right)$  directly. Instead, we use a low-rank approximation of the sensor-noise covariance matrix $R$, and perform the inversion in projected space:
\begin{equation}
    (J^T\Sigma J + R)^{-1} \approx U(U^TJ^T\Sigma JU + S_R)^{-1}U^T
\end{equation}
where $S_R$ contains the top-most singular values of $R$. For simplicity, the state prediction covariance $Q$ was approximated as $Q=\gamma I_{n\times n} $, where $\gamma = 10^{-6}$. A first-order transition model is used, where next-state joint-value estimates are defined as $ x'_{t+1} = x_{t} + \left(x_{t} - x_{t-1}\right) \Delta t$ for a fixed time-step $\Delta t$. We provide an initial prior over the state values with identical covariance, and an arbitrary mean-offset from ground-truth. 

Although it is possible to define an EKF in this manner, this approach is only expected to perform accurately for sufficiently smooth functions. The non-parametric component described in Section \ref{fwd_model} prevents the current forward model from having this property, and, the EKF derived for our forward model has limited accuracy in practice (see the experimental results in Section~\ref{sec_results} below). Therefore, instead of tracking with the forward model we learn an inverse sensor model, in analogy with inverse kinematics, to directly predict robot state from observations.


\subsection{An Inverse Sensor Model}
\label{inv_model}
\begin{figure}[h]
    \begin{center}
        \includegraphics[width=0.45\textwidth]{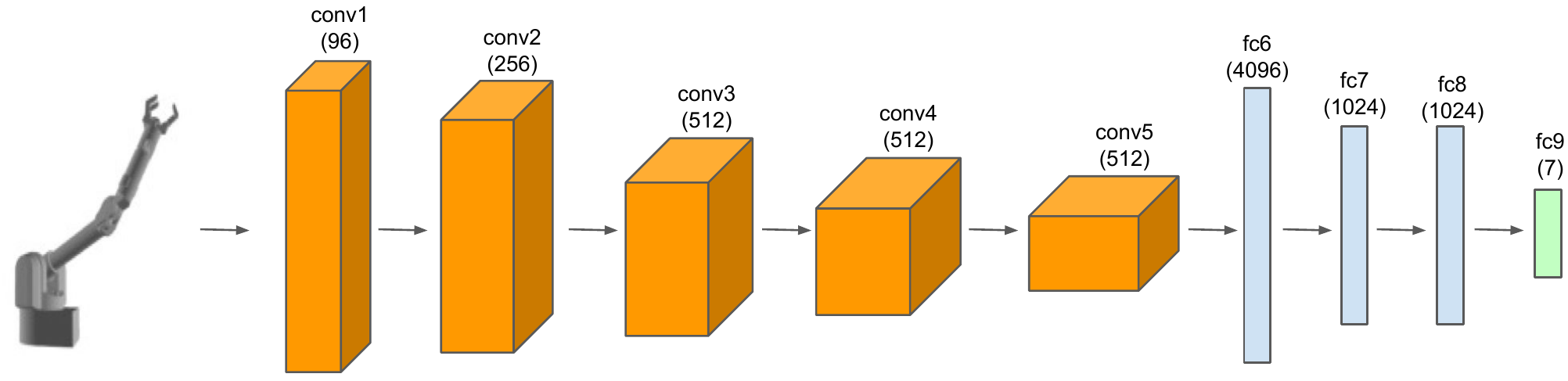} 
    \end{center}
    \caption{\small Inverse sensor model architecture.}\vspace{-4.0mm}
    \label{fig:arch_inverse}
\end{figure}

In order to infer the latent joint states $x_i$ from observed images $o_i$, we define a discriminative model $g_{fwd}$. Given the capacity of convolutional neural network models to perform regression on high-dimensional input data, we used a modified version of the VGG CNN-S network~\cite{chatfield2014return} containing 5 convolutional layers and 4 fully-connected layers (shown in Fig.\ref{fig:arch_inverse}). The model was trained on $256\times256\times3$ input images and corresponding joint labels, optimizing an $L_2$ loss on joint values and regularized with dropout layers.

\subsubsection{Tracking using the Inverse Sensor Model}
	The problem of tracking the state of a system can be generally defined as follows: given a history of observations $O_t = \left( o_i \right)_{i=t-T}^t$, we would like to estimate the corresponding sequence of states $X_t= \left( x_i \right)_{i=t-T}^t$. In the current application, we propose to use the inverse model described in Section \ref{inv_model} to infer each state value $x_i$ directly from $o_i$, and independently of other states ($x_{j\neq i} \in X_t$) and observations ($o_{j\neq i} \in O_t $). This assumes that the state of the system is fully-observable in a given image frame, and that the observations $o_i$ are therefore free of occlusion. 

%% file: results.tex
\begin{figure*}[htb]
    \begin{center}
        \begin{subfigure}[b]{0.3\textwidth}
        \includegraphics[width=\textwidth,keepaspectratio]{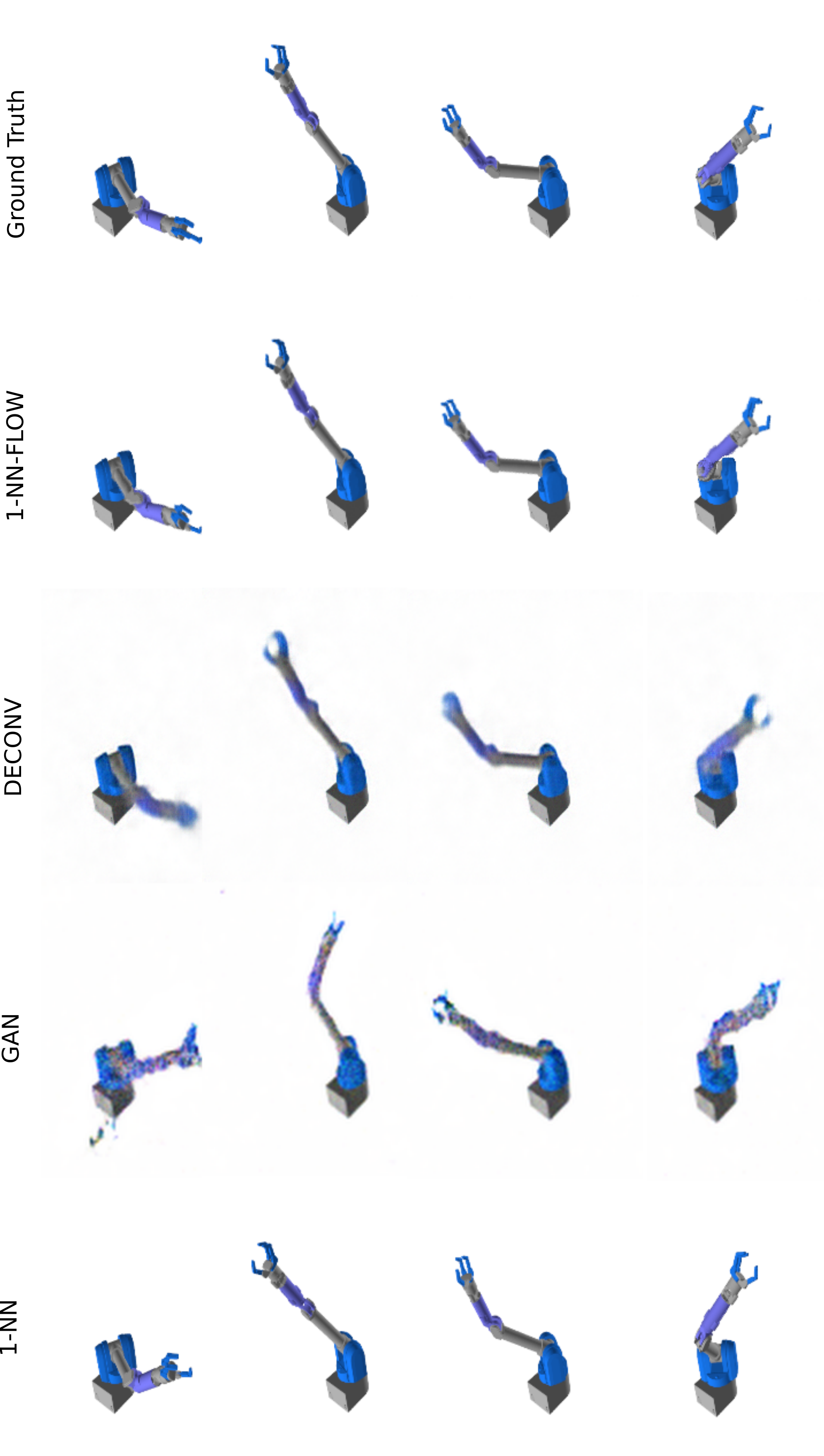}
        \caption{\small WAM 4-dof}
        \label{fig:sim_wam_4-dof}
        \end{subfigure}\qquad\qquad
         \begin{subfigure}[b]{0.42\textwidth}
        \includegraphics[width=\textwidth,keepaspectratio]{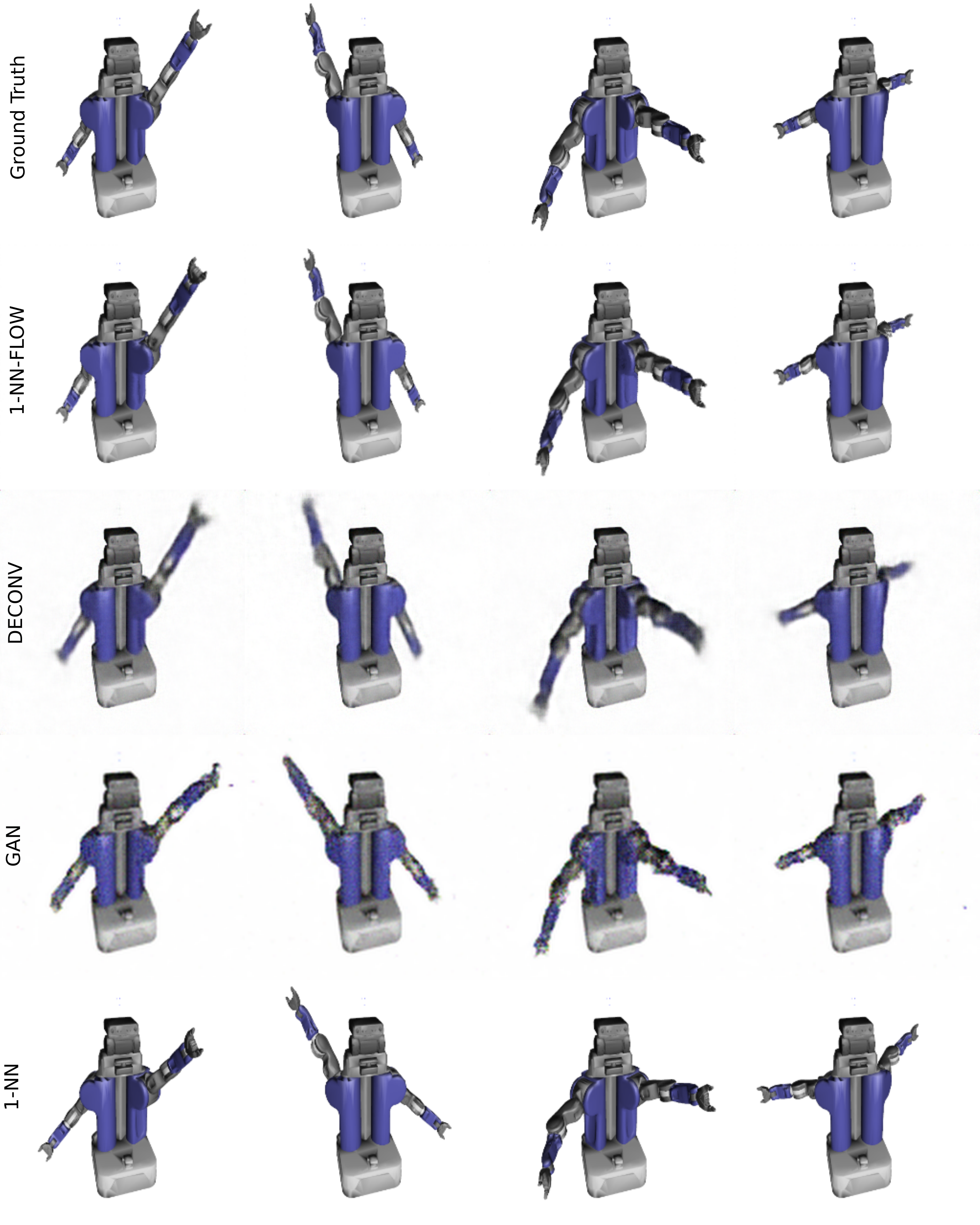}
        \caption{\small PR2 6-dof}
        \label{fig:sim_pr2_6-dof}
        \end{subfigure}
    \end{center}
    \caption{\small Examples of generated images for randomly-sampled input joint values, which were not encountered during training. Each row corresponds to the following (top-to-bottom):  ground-truth images, 1-NN-FLOW predicted output, DECONV baseline, and GAN baseline. Differences in robot-pose arise for GAN predictions due to noise injection necessary for joint-conditioned training.} \vspace{-3.0mm}
    \label{fig:forward_eval_sim}
\end{figure*}
\begin{figure}[htb]
    \begin{center}
		\includegraphics[width=\linewidth]{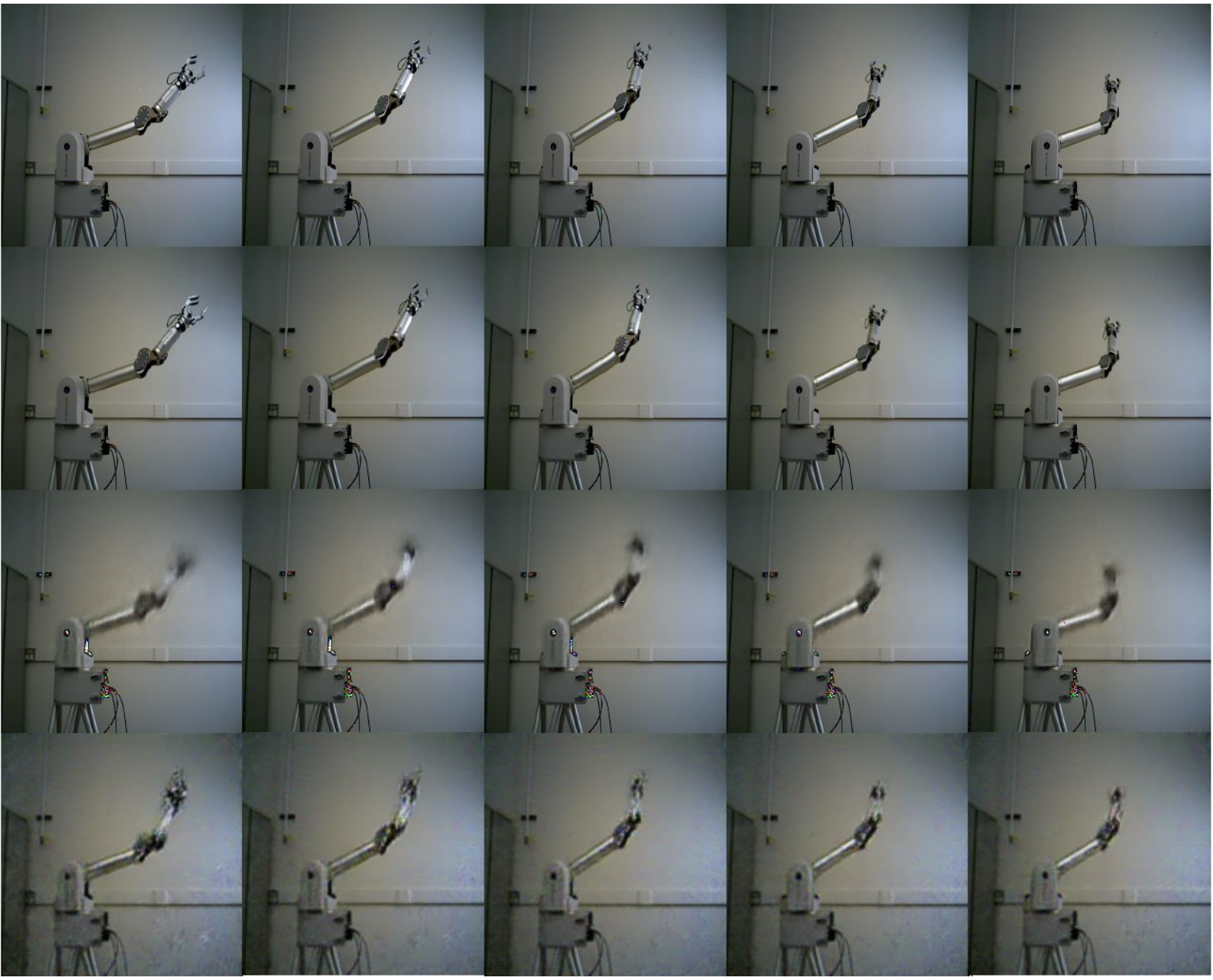}
    \end{center}
    \caption{\small Sequence of generated images for a test trajectory, where the top-to-bottom row correspondence is as follows:  ground-truth images, 1-NN-FLOW predicted output, DECONV baseline, and GAN baseline. Times indicated for $t=1,20,40,...,80$, from left to right, respectively.} \vspace{-3.0mm}
    \label{fig:forward_eval}
\end{figure}

\section{Experimental Results}
\vspace{-1.0mm}
\label{sec_results}
\subsection{Datasets}

\subsubsection{Simulated}

Images of robot-pose and corresponding joint values were captured for a Barrett WAM robot manipulator modeled in simulation using DART \cite{dart}. These were taken from a different viewpoint compared to those of the physical system, and the first four joints were sampled randomly. With a truly uniform sampling distribution, less data was required for training (20,000 for the training set and 10,000 for nearest-neighbor set). In addition, a simulation dataset of identical size was captured for a PR2 dual-manipulator robot, with random sampling of six degrees-of-freedom (first three joints in each arm). 

\subsubsection{Real-world}

Experiments were conducted using a Barrett WAM manipulator, a cable-actuated robotic arm. Data was captured from raw joint-encoder traces and a statically-positioned camera, collecting $640\times480$ RGB frames at 30 fps  using a PrimeSense camera (the infra-red depth modality was not used for this study). Due to non-linear effects arising from joint flexibility and cable stretch, large joint velocities and accelerations induce discrepancies between recorded joint values and actual positions seen in the images \cite{boots2014}. 
In order to mitigate aliasing, the joint velocities were kept under $10^{\circ}/s$. This and other practical constraints imposed limitations on obtaining an adequate sampling density of the four-dimensional joint space. As such, the training data was collected while executing randomly generated trajectories using only the first four joints (trajectories were made linear for simplicity). A total of 500 trajectories were executed, resulting in 225,000 captured camera frames and corresponding joint values. 50,000 data points were reserved for the nearest-neighbor data-set, and the remaining for training data. Test data was collected for arbitrary joint trajectories with varying velocities and accelerations (without concern for joint flexibility and stretch).

Dense sampling along sparsely-distributed trajectories results in non-uniformly sampled data. Simply constructing a nearest-neighbor dataset from randomly sampling collected data results in clusters of neighbors with high intra-cluster similarity. Picking $k$-nearest neighbors results in many reference images providing nearly identical viewpoints of the robot arm, which  reduces the benefit of using multiple neighbors to begin with. 
To introduce more dissimilarity (and variation) into $k$-neighbor selection, it was ensured that no two nearest neighbors originated from the same trajectory executed during data collection. 

\subsection{Forward Sensor Model Evaluation}

We first examine the predicted observations generated by the proposed forward sensor model ($k$-NN-FLOW). 
This includes three different variations of the proposed model, constructed for $k=1,2,5$ nearest neighbors. A deconvolutional network (DECONV) similar to that used in \cite{dosovitskiy2016} is selected as a baseline, with the absence of a branch for predicting segmentation masks (as these are not readily available from RGB data). Given recent successes of generative adversarial networks (GANs) for image generation, we also include a GAN model based on \cite{mirza2014conditional,dcgan}. GAN models can be conditioned for controlled output \cite{gauthier2014conditional}. Instead of conditioning on discrete labels, we condition on a continuous signal provided by the joint angles. We stabilize the training by adding low variance noise to the input signal . Finally, we include results produced by a simple 1-Nearest Neighbor implementation. 

Quantitative results for prediction accuracy are shown in Tables \ref{table:fwd-model-sim}  and \ref{table:fwd-model}, where it is apparent that both 1-NN-FLOW and 2-NN-FLOW models outperform the DECONV, GAN and 1-NN baselines in both mean $L_1$ and RMS pixel error. The 5-NN-FLOW predictions are similar in appearance to the real-world predictions produced by both the DECONV and GAN networks, but outperform them on the simulated datasets. All models outperform simple 1-NN selection.

Qualitative comparisons between the ground-truth, forward-sensor predictions, DECONV and GAN outputs are depicted in Fig.\ref{fig:forward_eval_sim} for randomly selected joint input values in simulation. For the physical system, prediction results are shown in Fig.\ref{fig:forward_eval} for a sequence of state-input values at various times on a pair of real test trajectories.


Detailed texture and features have been preserved in the $k$-NN-FLOW predictions, and the generated robot poses closely match the ground truth images. Both DECONV and GAN outputs suffer from blurred reconstructions, as expected, and do not manage to render certain components of the robot arm (such as the end-effector).

\begin{table}[!htbp]
\centering
\begin{tabular}{c|*2c*2c}
\toprule
	{}   		       & \multicolumn{2}{c}{WAM (4-dof)}        & \multicolumn{2}{c}{PR2 (6-dof)} \\ 
    Model			   & Mean $L_1$        &  RMS 	          & Mean $L_1$   &  RMS   \\ \midrule 
    5-NN-FLOW (ours)   &   0.00255         &   0.02319             &   0.00878        &   0.04546  \\ 
    2-NN-FLOW (ours)   &   0.00222         &   0.02171        &   0.00867        &   0.04843  \\  
    1-NN-FLOW (ours)   & \textbf{0.00183}  & \textbf{0.01905} & \textbf{0.00616} &  \textbf{0.03868}  \\  \hline 
    DECONV (baseline)  &   0.00726         &   0.02871        &   0.01567        &   0.04566  \\ 
    GAN (baseline)     &   0.03552         &   0.04262        &   0.12373        &   0.08151  \\
    1-NN (baseline)    &   0.00838         &   0.06471        &   0.02617        &   0.12283  \\
\bottomrule
\end{tabular}
\caption{\small RGB prediction pixel-error results on simulated datasets for WAM (4-dof) and PR2 (6-dof) platforms. Shown are values for different $k$-NN-FLOW forward models ($k=1,2,5$ nearest-neighbors), with comparison to 1-nearest-neighbor, DECONV and GAN baselines. Raw pixels values are within \( [0,1]\).} \vspace{-3.0mm}
\label{table:fwd-model-sim}
\end{table}

\begin{table}[!htbp]
\centering
\begin{tabular}{c|*2c}
\toprule
	Model   		   & Mean $L_1$          &  RMS \\   \midrule
    5-NN-FLOW (ours)   &  0.01246            &  0.02852  \\ 
    2-NN-FLOW (ours)   &  \textbf{0.01084}   &  \textbf{0.02269} \\ 
    1-NN-FLOW (ours)   &  0.01223            &  0.02510  \\ \hline 
    DECONV (baseline)  &  0.01303            &  0.02832   \\ 
    GAN (baseline)     &  0.01417            &  0.03017   \\ 
    1-NN (baseline)    &  0.01786            &  0.05210   \\
\bottomrule
\end{tabular}
\caption{\small RGB prediction pixel-error results on the real-world 4-dof WAM dataset. Shown are values for different $k$-NN-FLOW forward models ($k=1,2,5$ nearest-neighbors), with comparison to 1-nearest-neighbor, DECONV and GAN baselines. Raw pixels values are within \( [0,1]\).}\vspace{-3.0mm}
\label{table:fwd-model}
\end{table}

\subsection{Occlusion Detection}
\begin{figure}[h]
    \begin{center}
  \includegraphics[width=0.35\textwidth,height=.8\linewidth]{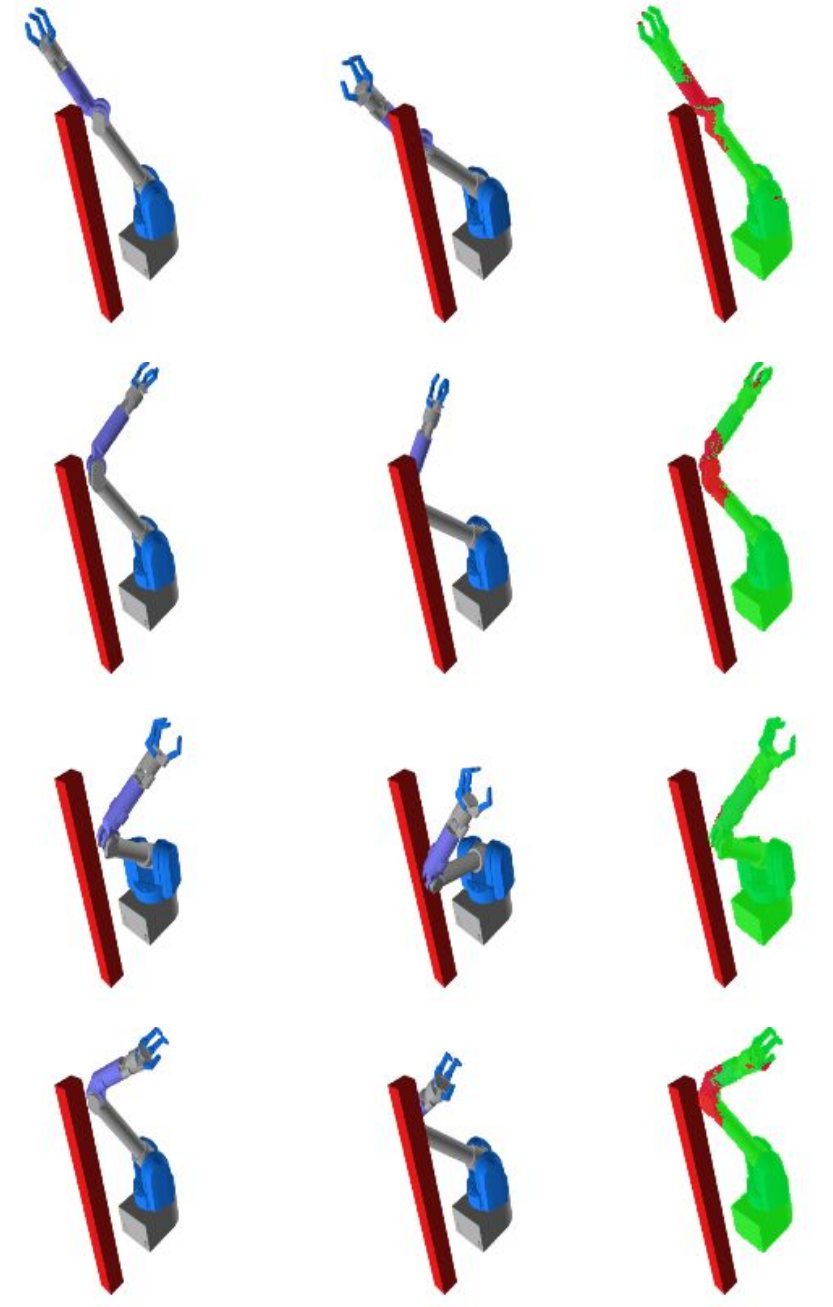} 
    \end{center}
    \vspace{-4.0mm}
    \caption{\small Occlusion Detection: Each row shows an example of occlusion detection. The first column is the first observation $o_1$ in which the arm is not occluded by the object. The goal is to predict which part of the arm would be occluded if it moves to the second state. Second columns show the second observation $o_2$. Note that the observations are shown for demonstration purposes, and the network only uses the corresponding joint values $x_1$ and $x_2$. The third column depicts the forward-backward symmetry check results on $o_1$: the red pixels shows the region that violates the forward-backward symmetry assumption (are occluded in the second state) and green regions will be visible in the second state.}\vspace{-4.0mm}
    \label{fig:occ_det}
\end{figure}

While collecting ground-truth optical flow values is infeasible in a real-world application, the core architecture shown in Figure~\ref{fig:arch_param} learns to predict the inverse (backward) optical flow fields from reference observation $o_r$ to the current observation $o$ without any direct supervision. The forward optical flow can also be generated by changing the order of the inputs to the network. While there are numerous applications that can use the optical flow information to improve their performance, in this section we show how the optical flow could be used for occlusion detection. We use the assumption that forward and backward optical flow are symmetric except for the occluded parts~\cite{alvarez2007symmetrical,ince2008occlusion}. Given two states $x_1$ and $x_2$, we compute the forward and backward optical flow. We then find the robot-pixels in the first image with symmetric forward and backward optical flow. Points that violate this property are assumed to be occluded in the second image. Figure~\ref{fig:occ_det} illustrates a few qualitative examples that were generated using this method. Predicting occlusions by projecting from configuration space into image frames allows such perceptual constraints to be used in defining desirable trajectories (for example, maintaining end-effector visibility), and could be used in conjunction with a planning framework. We leave further development and testing on real-world scenarios for future work. 
\subsection{Tracking Task Evaluation}
\vspace{-1mm}

We use the inverse model described in Section \ref{inv_model} to track the 4-dof robot joint positions from images. The performance is compared to tracking with an EKF formulated using the forward model (as described in Section \ref{fwd_model}) and an EKF based on the DECONV model. The latter uses a Jacobian calculated in the same manner as described in Section \ref{fwd_model}.

\begin{figure*}[htb]
  \centering
  \begin{subfigure}[b]{0.3\textwidth}
  	\centering
    \includegraphics[width=\textwidth] {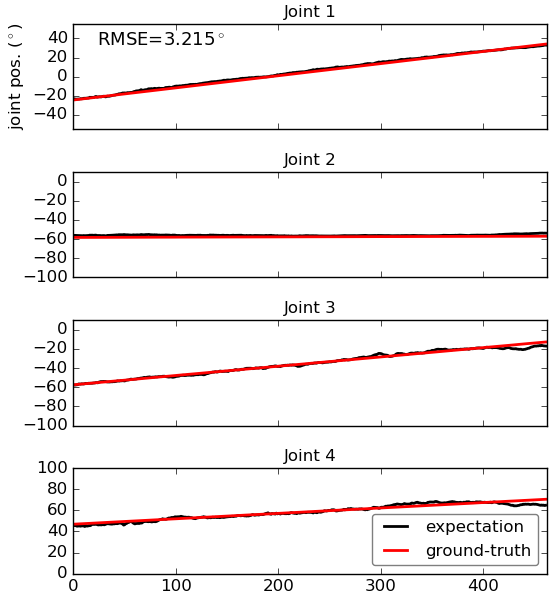} 
  	\caption{\small }
  \end{subfigure}
  \begin{subfigure}[b]{0.3\textwidth}
  	\centering
    \includegraphics[width=\textwidth] {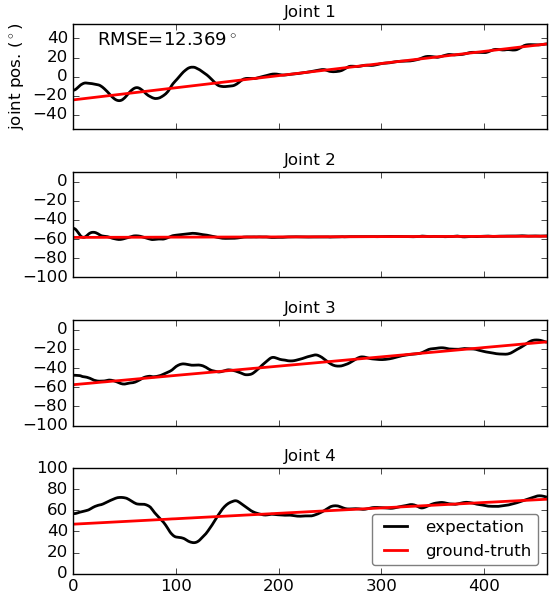} 
  	\caption{\small }
  \end{subfigure}
  \begin{subfigure}[b]{0.3\textwidth}
  	\centering
    \includegraphics[width=\textwidth] {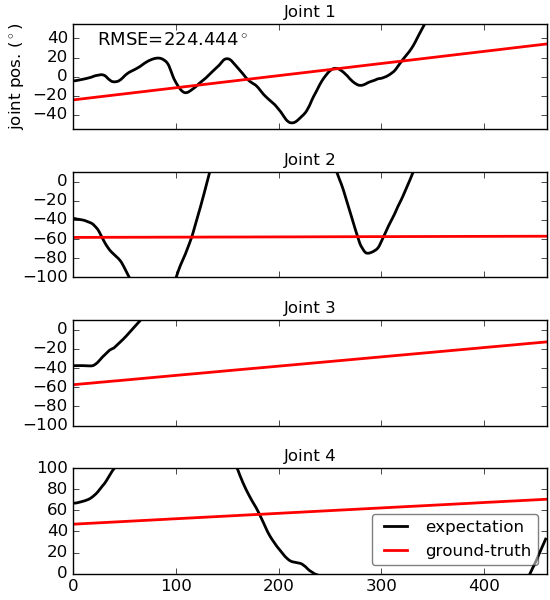} 
  	\caption{}
    \label{fig:fail_20deg}
  \end{subfigure}
  \caption{\small Comparison between the learned inverse sensor model (Figure~\ref{fig:arch_inverse}) and an EKF using the 2-NN-FLOW forward sensor model. Each row corresponds to a single joint evolving over 450 frames. The red line is the ground truth joint configuration, the black line is the estimated state. RMSE scores are shown. (a) The inverse model can robustly and accurately predict the state from an arbitrary image and unknown start state. (b) Tracking using an EKF and the learned DECONV model starting from a 10-degree offset and (c) 20-degree offset. The EKF works better when the state is already accurately tracked, but in general is much less robust and accurate than the learned inverse model.}
  \label{fig:tracking_2NN}
\end{figure*}

\begin{figure*}[htb]
  \centering
  \begin{subfigure}[b]{0.3\textwidth}
  	\centering
    \includegraphics[width=\textwidth] {./img_plots/results/conv_net_test_plot.png} 
  	\caption{}
  \end{subfigure}
  \begin{subfigure}[b]{0.3\textwidth}
  	\centering
    \includegraphics[width=\textwidth] {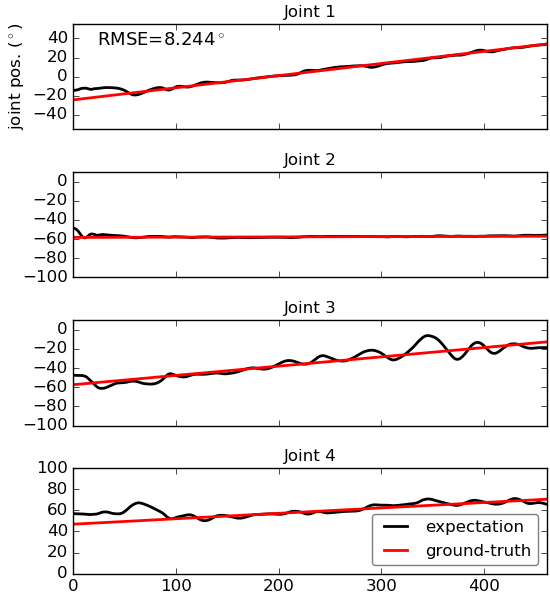} 
  	\caption{}
  \end{subfigure}
  \begin{subfigure}[b]{0.3\textwidth}
  	\centering
    \includegraphics[width=\textwidth] {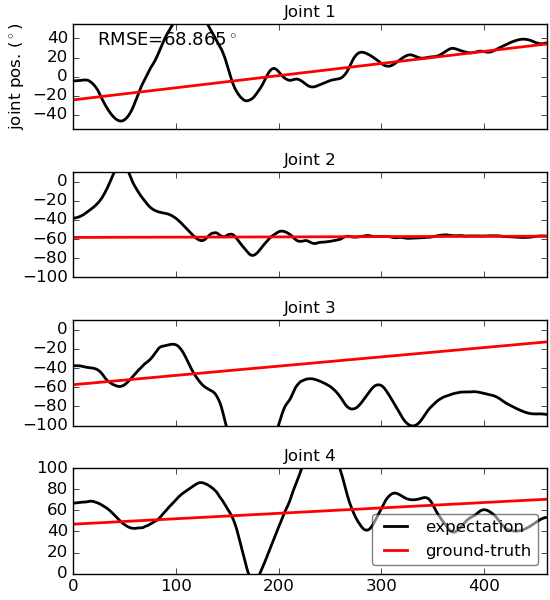} 
  	\caption{}
    \label{fig:fail_20deg_deconv}
  \end{subfigure}
  \caption{\small Tracking results and RMSE scores for DECONV EKF (same context as Figure \ref{fig:tracking_2NN}). The DECONV EKF works better when the state is already accurately tracked, but in general is much less robust and accurate than the learned inverse model. }\vspace{-3.0mm}
  \label{fig:tracking_DECONV}
\end{figure*}

Examples of tracking accuracy for a single test trajectory are shown in Figures \ref{fig:tracking_2NN} and \ref{fig:tracking_DECONV}. Here, joint state estimates from the inverse model are compared to those provided by a 2-NN-FLOW EKF and a DECONV EKF. Both EKF models are initialized with different mean-offsets ($x_0=10^\circ$ and $x_0=20^\circ$), to demonstrate approximate knowledge of the starting state at $t=0$.

\begin{figure}[htb]
  \centering
        \includegraphics[width=0.35\textwidth]{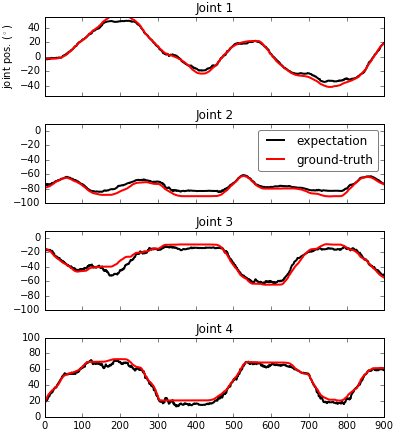} 
  \caption{\small Inferred joint values from a sequence of images using inverse sensor model for an arbitrary trajectory.}\vspace{-3.0mm}
  \label{fig:conv_nonlinear}
\end{figure}

The results indicate that both EKF trackers are able to converge the state estimate to the true trajectory over time, given a favorable ($10^\circ$ offset) initial prior. RMSE tracking errors demonstrate similar tracking performance for this initialization.
Failure cases for 2-NN-FLOW and DECONV EKF models are shown in Figures \ref{fig:fail_20deg} and \ref{fig:fail_20deg_deconv}. Here, the state is initialized with a $20^\circ$ offset, leading to instability in both trackers. Although performance has drastically deteriorated in both cases, it is worth noting that the DECONV EKF manages to recover reasonable joint state estimates for the first two joints. As mentioned in Section \ref{fwd_tracking}, EKF performance is dependent on the smoothness of the function, and may be a contributing factor to tracking robustness.
The stability of the tracking is highly dependent on the quality of the generated images. Discontinuities arise from the non-parametric component of the $k$-NN-FLOW model, as selected nearest-neighbors may abruptly change during a tracked trajectory. Although this effect is mitigated by using a softmax mask for weighting of nearest-neighbor-flow outputs, these sudden jumps in the value of the Jacobian and residual terms can lead to aberrations in the tracking performance.
It is apparent that the inverse sensor model is considerably more suited to the tracking task in this domain. The tracker requires no initial prior, and finds an optimum in a single forward pass (as opposed to the iterative optimization performed by the EKF). 
The comparatively high robustness of the inverse model in tracking is further demonstrated by estimating the state of an arbitrary nonlinear test trajectory shown in Figure \ref{fig:conv_nonlinear}. No latent dynamics model is assumed here, and state estimates are produced independently given a currently observed frame.

%% file: conclusions.tex
\section{Conclusions}
\vspace{-1.0mm}
We present a framework for tracking and prediction consisting of separate inverse and forward models, relating state to perceptual space, for purposes of state estimation and observation generation respectively. We propose a novel approach that combines the strengths of nearest neighbors and neural networks to generate high-quality predictions of never-before-seen images. In both a quantitative and qualitative sense, this generative network produces improved results over the DECONV, GAN and 1-NN baselines. For state-estimation and tracking, the generative observation model can be used in an EKF-based framework to perform probabilistic inference on the underlying latent state, and track the manipulator state over a simple trajectory. However, we show that learning a convolutional neural network as a forward model results in better performance in practice. We perform several experiments on a real robotic system, validating our approach and showing that our forward model is quantitatively and qualitatively state-of-the-art.
